\documentclass[letterpaper, 10 pt, conference]{ieeeconf}
\IEEEoverridecommandlockouts  
\usepackage{graphicx}  
\usepackage[caption=false,font=footnotesize]{subfig}
\usepackage[dvipsnames,table]{xcolor}  
\usepackage{amsmath}
\usepackage{amssymb}

\usepackage{amsthm} 

\newtheorem*{remark}{Remark}
\usepackage{mathtools}  
\usepackage{siunitx}  
\usepackage{algorithm}  
\usepackage{algpseudocode}
\usepackage{booktabs}  
\usepackage{multirow}
\usepackage{hhline}
\usepackage[export]{adjustbox}  

\usepackage{etoolbox}
\makeatletter

\patchcmd{\@makecaption}
  {\\}
  {.\ }
  {}
  {}

\makeatother

\definecolor{lightgreen}{rgb}{0.373,0.812,0.592}
\definecolor{lightyellow}{rgb}{0.98,0.953,0.702}

\newcommand{\Real}{\mathbb{R}}
\newcommand{\Vb}{$V_{\alpha}$}
\newcommand{\Vbset}{$\{V_{\alpha}\}$}
\newcommand{\Vbw}{$V_{\alpha}^{\omega}$}
\newcommand{\Vbwplus}{$V_{\alpha}^{\omega+1}$}
\newcommand{\Vbwset}{$\{V_{\alpha}^{\omega}\}$}
\newcommand{\Vbwplusset}{$\{V_{\alpha}^{\omega+1}\}$}
\newcommand{\GCBE}[1]{$G_C^{\omega#1}(\mathcal{B}, \mathcal{E})$}
\newcommand{\relsubopt}{$\Delta_{rel}$}

\newcommand{\Bconn}{$\mathcal{B}_{\mathrm{conn}}$}
\newcommand{\Ebw}{$E_{\alpha}^{\omega}$}
\newcommand{\Ebwout}{$E_{\alpha,out}^{\omega + 1 \setminus \omega}$}
\newcommand{\Ebwin}{$E_{\alpha,in}^{\omega + 1 \setminus \omega}$}

\def\amm{AMM-PGO$^{\#}$}
\def\rbcdpp{RBCD{\small++}}
\def\name{ROBO}
\newcommand{\namew}[1]{ROBO-#1}

\DeclareMathOperator{\SO}{SO}
\DeclareMathOperator{\SE}{SE}
\DeclareMathOperator{\Log}{Log}

\newcommand{\Rtilde}{\widetilde{R}}
\newcommand{\ttilde}{\widetilde{t}}
\newcommand{\Ttilde}{\widetilde{T}}

\makeatletter
\def\bstctlcite{\@ifnextchar[{\@bstctlcite}{\@bstctlcite[@auxout]}}
\def\@bstctlcite[#1]#2{\@bsphack
\@for\@citeb:=#2\do{%
\edef\@citeb{\expandafter\@firstofone\@citeb}%
\if@filesw\immediate\write\csname #1\endcsname{\string\citation{\@citeb}}\fi}%
\@esphack}
\makeatother

\title{\Large
Overlapping Domain Decomposition for Distributed Pose Graph Optimization
}

\author{Aneesa Sonawalla$^{1,2}$, Yulun Tian$^{3}$, and Jonathan P. How$^{1}$
\thanks{This material is based upon work supported by the National Science Foundation Graduate Research Fellowship Program under Grant No. 2141064, the Draper Scholars Program, and in part by ARL DCIST under Cooperative Agreement Number W911NF-172-0181.}
\thanks{$^{1}$Aneesa Sonawalla and Jonathan P. How are with the MIT Department of Aeronautics and Astronautics.
        {\tt \{aneesa, jhow\}@mit.edu}.}%
\thanks{$^{2}$Draper Scholar, The Charles Stark Draper Laboratory, Cambridge, MA.}%
\thanks{$^{3}$Yulun Tian is with the University of Michigan Robotics Department.
        {\tt yulunt@umich.edu}.}%
}

\algrenewcommand\algorithmicindent{0.5em}

\usepackage[noadjust,sort,compress]{cite}

\makeatletter
\let\NAT@parse\undefined
\makeatother

\usepackage[colorlinks=true,linkcolor=black,citecolor=black,urlcolor=blue]{hyperref}

\begin{document}

\bstctlcite{IEEEexample:BSTcontrol}

\maketitle
\thispagestyle{empty}
\pagestyle{empty}

\begin{abstract}
We present \name{} (Riemannian Overlapping Block Optimization), a distributed and parallel approach to multi-robot pose graph optimization (PGO) based on the idea of overlapping domain decomposition.
\name{} offers a middle ground between centralized and fully distributed solvers, where the amount of pose information shared between robots at each optimization iteration can be set according to the available communication resources.
Sharing additional pose information between neighboring robots effectively creates \textit{overlapping} optimization blocks in the underlying pose graph, which substantially reduces the number of iterations required to converge.
Through extensive experiments on benchmark PGO datasets, we demonstrate the applicability and feasibility of \name{} in different initialization scenarios, using various cost functions, and under different communication regimes.
We also analyze the tradeoff between the increased communication and local computation required by \name's overlapping blocks and the resulting faster convergence.
We show that overlaps with an average inter-robot data cost of only 36~Kb per iteration can converge 3.1$\times$ faster in terms of iterations than state-of-the-art distributed PGO approaches.
Furthermore, we develop an asynchronous variant of \name{} that is robust to network delays and suitable for real-world robotic applications.

\end{abstract}

\section{Introduction}
In GPS-denied scenarios, autonomous operation of robot teams relies on robust and accurate trajectory estimation from noisy relative measurements.
\textit{Pose graph optimization} (PGO) forms the backbone of GPS-denied, long-term localization in state-of-the-art multi-agent systems \cite{ebadi_present_2024,chang_kimera-multi_2021,lajoie_door-slam_2020}.
Many systems adopt a centralized architecture \cite{chang_lamp_2022,schmuck_covins_2021,li_corb-slam_2018}, in which a central computation node receives measurements from all robots and solves the optimization using off-the-shelf, high-performance libraries such as GTSAM \cite{gtsam} and Ceres Solver \cite{Agarwal_Ceres_Solver_2022}.
To handle increasing scale, inter-robot privacy constraints, and real-world network connectivity, an alternative line of work takes a distributed approach to PGO \cite{chang_kimera-multi_2021,choudhary_distributed_2017,murai_robot_2024}.
There are several state-of-the-art distributed PGO (DPGO) methods that offer minimal inter-agent communication per iteration, resilience to network delays, limited trajectory sharing between robots, and even certifiable optimality \cite{tian_asynchronous_2020,tian_distributed_2021,fan_majorization_2024,mcgann_asynchronous_2024}.
Yet these methods suffer from poor convergence rates, often requiring hundreds or thousands of iterations to reach a high-precision solution that a centralized solver could reach in tens of iterations.
The limited inter-robot communication of these methods is appropriate for robotics applications using low-bandwidth wireless networks, such as the 0.25 Mbps achievable with the Digi XBee 3 Zigbee 3.0 module \cite{xbee}.
However, modern ground and air robots frequently carry modules with much higher throughput, like the 100 Mbps Silvus SL5200 radios \cite{silvus}.
In these cases, existing DPGO methods significantly under-utilize the available communication bandwidth.
In this work, we address this gap by investigating a middle ground design between the fully centralized and fully distributed extremes that can flexibly take advantage of the resources available.
Our method achieves substantially improved convergence rates via a controlled increase in communication at every optimization iteration, maintaining communication requirements well within the limits of modern networks.

To this end, we present \textbf{\name{}} (Riemannian Overlapping Block Optimization), a novel, parallel, and fast distributed PGO solver based on overlapping domain decomposition, a classical approach in numerical optimization~\cite{saad_iterative_2007}. 
We design an overlapping scheme for solving graph-structured optimization problems with on-manifold decision variables, drawing on prior works \cite{shin_decentralized_2020,shin_overlapping_2020,shin_exponential_2021} that have investigated similar decomposition approaches.
Multi-robot PGO naturally decomposes the larger PGO problem into blocks, or domains, corresponding to individual robots. 
Existing DPGO methods only require agents to communicate poses located at the boundary of their respective domains at every iteration.
With \name{}, robots selectively communicate additional information from the interiors of their domains.
This allows the original domains to inflate and overlap with each other, providing each robot more information on the global problem to accelerate convergence.
The degree of overlap, controlled by a hyperparameter $\omega$, allows our approach to trade off per-iteration communication and overall convergence rate.
Without overlap ($\omega=0$), our approach reduces to the previous distributed solver \cite{tian_distributed_2021} that operates on disjoint domains.
Meanwhile, with increasing overlap ($\omega \to \infty$), \name{} yields substantially faster convergence, eventually recovering centralized computation where every robot solves the global pose graph.
In practice, the degree of overlap applied can be set according to the available network bandwidth, allowing our method to be adapted to available resources to achieve faster convergence.
Further, the parallel nature of \name{} makes the method readily applicable under asynchronous communication.
Finally, while it might be expected that sharing more information improves estimation accuracy, \name's flexible design allows us to ask: \emph{how much} overlap is needed to improve convergence in common DPGO settings, and is the increased computation/communication tractable for robotics applications?
To the authors' knowledge, this is the first demonstration of convergence of this type of overlapping domain decomposition method for distributed, on-manifold optimization, as well as the first communication cost analysis for this type of approach.
In summary, our contributions are:
\begin{itemize}
    \item \textbf{\name{}} (Riemannian Overlapping Block Optimization), a \emph{fully parallel} PGO method that leverages overlapping domain decomposition to converge in substantially fewer iterations than state-of-the-art approaches;
    \item An analysis of the convergence improvement and communication cost trade-off when increasing overlap;
    \item An asynchronous variant of \name{} that is robust to network delays common in real-world applications;
    \item Extensive experiments on both synthetic and real-world benchmark datasets demonstrating \name's adaptability to different initialization schemes, optimization formulations, and communication paradigms.
\end{itemize}

\section{Related Work}\label{sec:related-work}

\subsection{Multi-Agent PGO}

Multi-agent PGO involves solving a nonlinear, non-convex optimization for multiple robots' poses given a set of noisy relative inter- and intra-robot measurements. This optimization can be solved in a centralized, decentralized, or distributed fashion. Fast, mature solvers exist for centralized multi-robot PGO \cite{andersson_c-sam_2008,bailey_decentralised_2011,zhang_mr-isam2_2021}, but centralized computation can become untenable as the size of the robot team or operational area scales up. Decentralized methods \cite{liu_slideslam_2024,lajoie_swarm-slam_2024,cunningham_ddf-sam_2010,cunningham_ddf-sam_2013} remove the need for a central server, but can suffer under communication or local computation constraints. This work focuses on distributed methods, which enable highly scalable multi-robot localization under real-world constraints \cite{halsted_survey_2021}.

\subsection{Distributed PGO}

Distributed PGO (DPGO) has been a problem of interest for many years, particularly over the past decade.
Choudhary et al.\ introduced a two-stage approach in DGS that uses the successive over-relaxation and the Jacobi over-relaxation \cite{choudhary_distributed_2017}.
Tian et al.\ developed DC2-PGO, which uses a distributed Riemannian staircase approach to solve a semidefinite relaxation of the PGO problem and can certify global optimality of the returned solution \cite{tian_distributed_2021}.
The authors also introduced Riemannian block coordinate descent (RBCD) and its accelerated variant \rbcdpp, which have provable convergence on any smooth optimization defined on the product of Riemannian manifolds \cite{tian_distributed_2021}.
Fan and Murphey introduced a provably convergent method that leverages an accelerated majorization minimization approach, along with an accelerated, distributed initialization scheme for multi-agent PGO \cite{fan_majorization_2020,fan_majorization_2024}.
Recently, Li et al.~\cite{li_distributed_2024} improved the acceleration of RBCD/\rbcdpp{} with IRBCD and further proposed a load-balanced graph partitioning approach in constructing the distributed optimization.

All of the above methods require synchronization across agents; other DPGO approaches exist that do not. Cunningham et al.\ introduced DDF-SAM, a decentralized system in which agents share reduced versions of their local graphs obtained from Gaussian elimination \cite{cunningham_ddf-sam_2010,cunningham_ddf-sam_2013}. Tian et al.\ developed ASAPP, a distributed and asynchronous PGO approach that remains provably convergent under bounded network delay \cite{tian_asynchronous_2020}. Murai et al.\ proposed a method based on Gaussian belief propagation that is fully asynchronous, distributed, and robust to communication delays and dropouts \cite{murai_robot_2024}. McGann et al.\ took an approach based on consensus alternating direction method of multipliers (C-ADMM) to develop MESA, which also empirically converges under communication limitations \cite{mcgann_asynchronous_2024}.

While the above methods have all made significant advances in DPGO in terms of convergence rate, robustness, and guaranteed convergence, the current state-of-the-art still requires many iterations to reach a high-precision solution that a centralized solver could reach in tens of iterations or fewer. These methods most often assume a maximally restrictive communication scheme in the interest of privacy constraints and limited network bandwidth, meaning they cannot take advantage of higher-throughput networks increasingly available in modern robotics applications.

\subsection{Overlapping Domain Decomposition}

Overlapping domain decomposition, also known as the overlapping Schwarz method, has historically been applied to solve large, sparse linear systems as a ``divide-and-conquer'' strategy \cite{saad_iterative_2007}. Shin et al.\ proved the convergence in both synchronous and asynchronous settings of an overlapping domain decomposition approach to graph-structured optimization problems that can be expressed as an unconstrained, convex quadratic program \cite{shin_decentralized_2020}.
A subsequent work extended this result to constrained quadratic programs in \cite{shin_overlapping_2020}. The basis for the authors' convergence proofs is that the sensitivity of the solution at any node in the graph to a change in the solution at another node decays exponentially with the distance between those nodes \cite{shin_exponential_2021}. A similar property was shown in \cite{hamam_streaming_2022} for optimization problems with locally coupled costs. Shin et al.\ showed this sensitivity property holds for graph-structured nonlinear programs whose objective functions meet certain conditions \cite{shin_exponential_2021}, while Na et al.\ showed it holds for nonlinear optimal control problems \cite{na_convergence_2022}. While these works developed Schwarz-like schemes for a variety of graph-structured optimization problems, to the best of our knowledge, the literature does not include an exploration of convergence for the case of graph-structured optimization with on-manifold decision variables like PGO. Additionally, there is no analysis of the tradeoff between communication cost and convergence rate that is essential to consider in real-world robotics applications.

\section{Problem Formulation}\label{sec:problem}

In this section, we formally introduce the distributed PGO problem and relevant mathematical preliminaries. 
Our formulation is compatible with multiple commonly used objective function definitions.
We also consider several communication scenarios relevant to real-world DPGO applications. 

\subsection{Preliminaries}

The \textit{special orthogonal group} is denoted as $\SO(d) \triangleq \{R \in \Real^{d \times d} \, | \, R^\top R = I_d, \, \det(R) = 1 \}$, where $I_d$ is the $d$-dimensional identity matrix and $d=2$ or $d=3$ for 2D or 3D poses. The \textit{special Euclidean group} is denoted by $\SE(d) \triangleq \SO(d) \ltimes \Real^d$. The set of $n$ poses to be estimated are $T_1, \ldots, T_n$, where $T_i = (R_i, t_i)$ includes rotation $R_i \in \SO(d)$ and translation $t_i \in \Real^d$.  We consider noisy, full-rank relative measurements between poses also consisting of a rotation and translation, with the measurement between poses $i$ and $j$ given by $\Ttilde_{ij} = (\Rtilde_{ij}, \ttilde_{ij})$, $\Rtilde_{ij} \in \SO(d)$ and $\ttilde_{ij} \in \Real^d$.

\subsection{Pose Graph Optimization}

We can model the PGO problem as a directed, connected graph $G = (V, E)$ in which the set of nodes $V$ corresponds to pose variables $\{T_i\}_i^{|V|}$ to be estimated, and the set of edges $E$ correspond to pairwise relative measurements $\{\Ttilde_{ij} \}$ from $i$ to $j$.
$|V|$ is the size of the set $V$, equivalent to the number of poses to be estimated.
The PGO problem is formulated as
\begin{equation}\label{eqn:pgo}
\begin{split}
    &\min_{ \{T_i\}_{i \in V}} \dfrac{1}{2} \sum_{(i,j) \in E}
    \mathrm{dist}^2_{\Omega_{ij}}(T_i \Ttilde_{ij}, T_j), \\
    &\mathrm{s.t.} \quad R_i \in \SO(d), \: t_i \in \Real^d \quad \forall \; T_i = (R_i, t_i), i \in V.
\end{split}
\end{equation}
In \eqref{eqn:pgo},
$\mathrm{dist}^2_{\Omega_{ij}}: \SE(d) \times \SE(d) \to \Real_{\geq 0}$ denotes the squared distance between two poses weighted by an information matrix $\Omega_{ij}$.
Following \cite{rosen_se_sync_2019,tian_distributed_2021,tian_asynchronous_2020}, we assume the translation and rotation noise are independent and isotropic. 
This means that each $\Omega_{ij}$ can be fully described by two scalars $\kappa_{ij}, \tau_{ij} > 0$ corresponding to the precisions of the rotation and translation measurements, respectively.
Using the chordal distance metric for rotations, this formulation recovers the same objective function used by SE-Sync \cite{rosen_se_sync_2019} and DC2-PGO \cite{tian_distributed_2021}, 
where 
$\mathrm{dist}^2_{\Omega_{ij}}(T_i \Ttilde_{ij}, T_j) = \kappa_{ij}|| R_j - R_i \Rtilde_{ij}||_F^2 + \tau_{ij}|| t_j - t_i - R_i \ttilde_{ij}||_2^2$.
We also consider an alternative objective that uses the geodesic distance for rotations.
In this case, 
$\mathrm{dist}^2_{\Omega_{ij}}(T_i \Ttilde_{ij}, T_j) = \kappa_{ij}|| \Log(\Rtilde_{ij}^T R_i^T R_j)||_2^2 + \tau_{ij}|| t_j - t_i - R_i \ttilde_{ij}||_2^2$.
The logarithm map $\Log(\cdot)$ converts a rotation matrix to its angle-axis representation \cite{hartley_rotation_2013}.

For DPGO, a team of $N$ robots collaboratively solves \eqref{eqn:pgo} without a centralized node.
Each robot solves a subset of poses in $V$.
We consider each robot's trajectory, or its ``owned'' poses, as a subgraph, providing a natural partitioning of $G(V,E)$ into 
\textit{blocks}, which we denote by \Vb{} with $\alpha \in \mathcal{B} \triangleq \{1, 2, \ldots, N\}$.
Blocks are connected by inter-robot measurements to form the global, connected graph.
Figure~\ref{fig:example_graphs} shows an example of a three-robot graph with intra- and inter-robot measurements with different colors (blue, orange, green) highlighting $V_{\alpha}$, $V_{\beta}$, $V_{\gamma}$.

\section{Methods}\label{sec:methods}

In this section, we start by reviewing Riemannian block coordinate descent (RBCD) \cite{tian_distributed_2021} and describe how we extend RBCD to develop \name{}.
We then detail how our method is adapted for the synchronous, edgewise, and asynchronous communication settings drawn from prior works \cite{fan_majorization_2024,tian_distributed_2021,mcgann_asynchronous_2024,tian_asynchronous_2020}.

\subsection{\name{}: PGO with Overlapping Domain Decomposition}

Block coordinate descent (BCD) methods are well-suited to DPGO because of the natural decomposition of the global pose graph into blocks corresponding to individual robots.
RBCD was introduced in \cite{tian_distributed_2021} as a general BCD algorithm applicable to smooth optimization problems over the product of Riemannian manifolds.
This class of optimization problems includes the PGO problem in~\eqref{eqn:pgo}.
To collaboratively perform PGO using RBCD, each robot only needs to share the ``boundary'' poses from its own block that are connected to neighboring robots by an inter-robot measurement.
From a centralized view, the basic steps of an RBCD iteration are 1) select a block (i.e., robot) to optimize; 2) approximately solve the optimization over that block, holding all other nodes in the graph constant; and 3) carry over all other values in the global graph.

\begin{figure}[t]
    \centering
    \subfloat[]{
        \begin{minipage}[t][][b]{0.6\columnwidth}
            \centering
            \includegraphics[width=\textwidth, height=0.75\textheight, keepaspectratio]{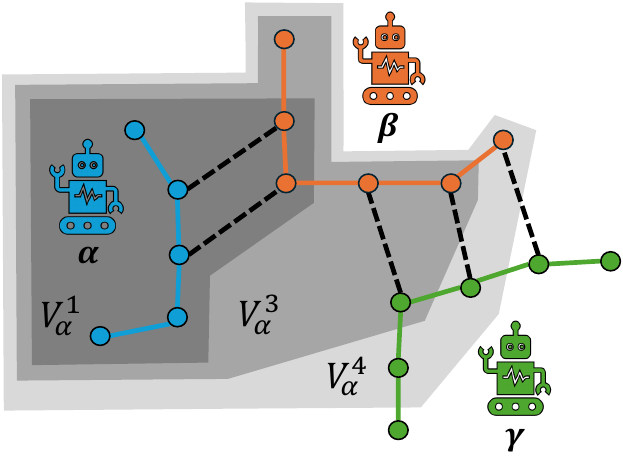}
            \vspace*{-0.3in}
            \label{fig:overlapped_pgo}
        \end{minipage}
    } \hfill
    \begin{minipage}[t][][b]{0.36\columnwidth}
        \centering
        \subfloat[$\omega = 1$]{\includegraphics[width=\textwidth, height=0.4\textheight, keepaspectratio]{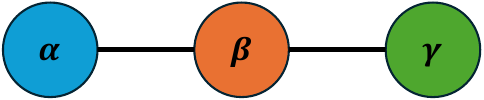}\label{fig:GC_overlap1}}
        \vspace{0.014\textheight}

        \subfloat[$\omega = 3, 4$]{\includegraphics[width=0.75\textwidth, height=0.2\textheight, keepaspectratio]{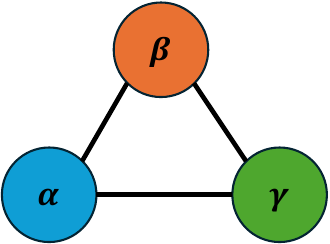}\label{fig:GC_overlap3}}
    \end{minipage}
    \caption{Example three-robot pose and communication graphs. (a) Nodes correspond to poses; node color indicates ``ownership'' of that pose. Solid lines denote odometry measurements, dashed lines are inter-robot loop closures. Overlapping blocks with $\omega = 1,3,4$ highlighted for robot $\alpha$. Robot subgraphs summarized in (b) and (c) to form the communication graph, \GCBE{}. (b) \GCBE{} for $\omega = 1$. (c) $\omega = 3$ or $4$ changes \GCBE{} due to inter-robot measurements, such that all robots are connected.}
    \label{fig:example_graphs} \vspace*{-0.27in}
\end{figure}

\name{} improves convergence rates by allowing more per-iteration communication than RBCD.
This is most useful when the primary goal is rapid convergence and resources afford higher inter-robot communication bandwidth, which is typical in many real-world settings with ground or air robots.
With \name{}, we extend the idea of RBCD to include \textit{overlapping} blocks constructed from overlapping domain decomposition \cite{saad_iterative_2007}.
Blocks no longer correspond just to one robot's poses (see Figure~\ref{fig:example_graphs}).
We further choose to optimize over all blocks in parallel, rather than alternating as in \cite{tian_distributed_2021}.

We first show how to create these overlapping blocks from the global graph.
Following \cite{shin_decentralized_2020}, we define the overlapping decomposition using the graph distance: let $d_G(i, j)$ be the number of edges in the shortest path between nodes $i, j \in V$. The distance between a node $i \notin X$ and a set of nodes $X$ is 
$d_G(i,X) = \min_{j \in X} d_G(i,j)$.
If $i \in X$, then $d_G(i,X) = 0$.
The overlapping set of nodes for robot $\alpha$ with overlap size $\omega$ is then defined as $V_{\alpha}^\omega \triangleq \{ i \in V \;|\; d_G(i,V_{\alpha}) \leq \omega \}$.
Note that \Vbw{} also includes robot $\alpha$'s nodes, i.e., \Vb $\subseteq$ \Vbw{}.

In \name, each robot $\alpha \in \mathcal{B}$ solves a subproblem of~\eqref{eqn:pgo} over its block \Vbw.
We denote the edges within the block \Vbw{}, or the ``interior'' edges of \Vbw{}, by $E_{\alpha}^{\omega} \triangleq \{(i,j) \in E \ |\ i,j \in V_{\alpha}^{\omega} \}$.
Note that $E_{\alpha}^{\omega}$ includes all edges---odometry and all loop closures---satisfying this definition.
The block \Vbw{} has boundary poses \Vbwplus $\setminus$ \Vbw{}.
When the remaining poses in $V \setminus$ \Vbw{} are fixed, these new boundary poses effectively form \emph{prior terms} that constrain the robot's local subproblem.
Let $\bar{T}_i$ denote a fixed boundary pose $i$.
Since $G$ is directed, the edges connecting the fixed poses to the block \Vbw{} can be either ``incoming'' or ``outgoing'' with respect to \Vbw.
Let the outgoing edges be $E_{\alpha, out}^{\omega+1 \setminus \omega} \triangleq \{ (i,j) \in E\  |\ i \in V_{\alpha}^{\omega},\ j \in  V_{\alpha}^{\omega+1} \setminus V_{\alpha}^{\omega} \}$, and the incoming edges be $E_{\alpha, in}^{\omega+1 \setminus \omega} \triangleq \{ (i,j) \in E\  |\ i \in  V_{\alpha}^{\omega+1} \setminus V_{\alpha}^{\omega}, \; j \in V_{\alpha}^{\omega} \}$.
With these definitions, the subproblem solved by robot $\alpha$ is:
\begin{equation}\label{eqn:pgo_subproblem}
\begin{split}
    &\min_{ \{T_i\}_ {i \in V_{\alpha}^{\omega}}} \dfrac{1}{2} \sum_{(i, j)  \in E^\omega_\alpha}
    \mathrm{dist}^2_{\Omega_{ij}}(T_i \Ttilde_{ij}, T_j) \\
    & ~~~~~~~~~~+ \dfrac{1}{2} \sum_{(i, j)  \in E^{\omega+1 \setminus\omega}_{\alpha, out}} \mathrm{dist}^2_{\Omega_{ij}}(T_i\Ttilde_{ij}, \bar{T}_j) \\
    & ~~~~~~~~~~+ \dfrac{1}{2} \sum_{(i, j)  \in E^{\omega+1 \setminus \omega}_{\alpha, in}} \mathrm{dist}^2_{\Omega_{ij}}(\bar{T}_i\Ttilde_{ij}, T_j) \\
    &\mathrm{s.t.} \quad R_i \in \SO(d), \: t_i \in \Real^d \quad \forall \; T_i = (R_i, t_i) \in V_{\alpha}^{\omega}.
\end{split}
\end{equation}

\begin{remark}[Communication graph]
As in RBCD \cite{tian_distributed_2021}, the subproblem blocks induce a robot-level communication graph.
To form the local subproblem, robot $\alpha$ needs \Vbw{} and \Ebw{}; the boundary poses, \Vbwplus $\setminus$ \Vbw{}; and the incoming and outgoing boundary edges, \Ebwin{} and \Ebwout.
This means robot $\alpha$ needs to form \Vbwplus{} to solve the subproblem for overlap $\omega$.
If we summarize \Vbwplus{} as a single node, we get a second graph showing the robot-level level connectivity needed for overlap $\omega$.
We call this graph the \emph{communication graph}, denoted \GCBE{}.
An edge exists in $\mathcal{E}$ between $\alpha \in \mathcal{B}$ and $\beta \in \mathcal{B}$ if and only if there is at least one pose $i \in$ \Vbwplus{} such that $i \in V_{\beta}$ or at least one pose $j \in V_{\beta}^{\omega+1}$ such that $j \in $ \Vb.
\end{remark}

Figure~\ref{fig:example_graphs} illustrates the above concepts. Figure~\ref{fig:overlapped_pgo} shows \Vbw{} for robot $\alpha$ in shades of gray for $w=1,3,4$. If we choose $\omega=3$, the three poses in $V_{\alpha}^4 \setminus V_{\alpha}^3$ are also needed by robot $\alpha$ to form the optimization in~\eqref{eqn:pgo_subproblem}.
Figures~\ref{fig:GC_overlap1} and~\ref{fig:GC_overlap3} show the communication graphs for several values of $\omega$
where the example pose graph from Figure~\ref{fig:overlapped_pgo} leads to different communication connectivity, depending on $\omega$.

\begin{figure}[t]
\begin{algorithm}[H]
\small
\caption{\textsc{\small \name{}: Overlapped Distributed PGO}}
\label{alg:robo}
\renewcommand{\algorithmicrequire}{\textbf{Input:}}
\renewcommand{\algorithmicensure}{\textbf{Output:}} 
\begin{algorithmic}[1]  
    \Require $G(V, E)$, $\omega$, \Vbset{}, convergence condition
    \Ensure $V$ at a local minimum of~\eqref{eqn:pgo}

    \State Determine \GCBE{} using $G(V, E)$, \Vbset{}
    \State Share initial estimates to form \Vbwplusset{} $\forall$ agents  $\alpha \in \mathcal{B}$ \label{alg:robo_init}
    \While{not converged}
    \For{$\alpha \in \mathcal{B}_{\mathrm{conn}}$ (in parallel)} \label{alg:robo_iterstart}
    \State \Vbw $\leftarrow$ approximately solve~\eqref{eqn:pgo_subproblem} \label{alg:robo_solve}
    \State Send $\{ v_{\alpha} \in V_{\alpha} \ |\ v_{\alpha} \in V_{\beta}^{\omega+1} \}$ to $\beta \in \mathcal{B}_{\mathrm{conn}}, \; (\alpha,\beta) \in \mathcal{E}$ \label{alg:robo_share}
    \State \Vbwplus $\setminus V_{\alpha} \leftarrow$ updates from $\beta \in \mathcal{B}_{\mathrm{conn}}, \; (\alpha,\beta) \in \mathcal{E}$ \label{alg:robo_update}
    \EndFor \label{alg:robo_iterstop}
    \EndWhile
\end{algorithmic}
\end{algorithm}
\vspace*{-0.35in}
\end{figure}

We can now detail the steps of our method (Algorithm~\ref{alg:robo}).
We assume the full graph $G(V,E)$, the overlap size $\omega$, and the graph partitioning to create \Vbset{} are known.
The measurement edges $E$ and the partitioning $V_{\alpha}$ are sufficient to determine the overlapped blocks, \Vbwset, and the blocks with boundary edges and poses included, \Vbwplusset{} for $\alpha \in \mathcal{B}$.
The communication graph \GCBE{} can then be determined using \Vbwplusset.
Different initialization schemes can be used in Step~\ref{alg:robo_init}; several options are explored in Section~\ref{sec:experiments}.

In each iteration of \name, agents solve their inflated subproblems in parallel, then project their local solutions onto their own domains \Vbset{} before sharing the required parts of their solutions with their neighbors.
This is reflected in Steps~\ref{alg:robo_iterstart}--\ref{alg:robo_iterstop}.
In general, we can have a different subset of robots \Bconn $\subseteq \mathcal{B}$ that perform local optimization at every iteration.
In Step~\ref{alg:robo_solve}, each agent in \Bconn{} approximately solves the subproblem in~\eqref{eqn:pgo_subproblem} locally over its block \Vbw{}, where \Vbwplus $\setminus$ \Vbw{} are used to construct the priors in~\eqref{eqn:pgo_subproblem}.
Here we use a single step of a Levenberg-Marquardt solver initialized with \Vbw{} from the last iteration to reach an approximate solution to \eqref{eqn:pgo_subproblem}. 
However, the method is not specific to which local solver is used.
In Step~\ref{alg:robo_share}, each agent in \Bconn{} shares the subset of its owned poses $V_{\alpha}$ that are also in its neighbors' overlapped blocks with each of those neighbors; that is, each agent $\alpha$ sends poses $v_{\alpha} \in V_{\alpha}$ to its neighbor $\beta$ if $v_{\alpha} \in V_{\beta}^{\omega + 1} \setminus V_{\beta}$.
We emphasize that only the overlapped portions of each agent's own and neighboring blocks are shared, not the entire set of poses $V_{\alpha}$.
Finally, in Step~\ref{alg:robo_update}, each agent in \Bconn{} updates its local graph \Vbwplus{} with the overlapped poses \Vbwplus $\setminus$ \Vb{} received in Step~\ref{alg:robo_share}.
The convergence condition may be set by local convergence criteria, by periodic synchronization across robots to determine global properties, or by a practical need such as a maximum allowable optimization time.

\subsection{Communication Schemes}
By default, the approach in Algorithm~\ref{alg:robo} is synchronized in that all agents in \Bconn{} must complete one round of parallel optimization and communication before proceeding to the next iteration. 
In the following, we discuss how Algorithm~\ref{alg:robo} can be executed in different communication scenarios.

\textbf{Synchronous \name{}.}
In the \textit{synchronous} communication paradigm, all agents are able to communicate with their neighbors at each iteration, and iterations are synchronized across the team. Thus, in Step~\ref{alg:robo_iterstart}, $\mathcal{B}_{\mathrm{conn}} = \mathcal{B}$. This paradigm leads to the most straightforward algorithm design, but it also places the most stringent requirements on the physical network, and is therefore best suited to applications with reliable network connections.

\textbf{Edgewise \name{}.}
As operating environment and team size scale, it is likely that not all agents will be in constant communication, meaning \Bconn{} in Step~\ref{alg:robo_iterstart} will be time-varying and in general a subset of $\mathcal{B}$.
To model this limited network connectivity, we consider an \textit{edgewise-only} scheme, inspired by \cite{mcgann_asynchronous_2024}.
We assess the performance of our method in the most restrictive edgewise case as in \cite{mcgann_asynchronous_2024}, wherein only two agents can communicate at every iteration.
In this regime, an agent may optimize using stale data from agents it has not communicated with recently.
We simulate this by randomly sampling a robot pair $(\alpha, \beta) \in \mathcal{E}$ at every iteration and setting \Bconn $= \{ \alpha, \beta \}$ for that iteration. The rest of Algorithm~\ref{alg:robo} remains the same.

\textbf{Asynchronous \name{}.}
Lastly, we consider the \textit{asynchronous} regime.
While the edgewise scenario relaxes communication requirements across the team, it still assumes alternating communication and optimization steps.
In contrast, the asynchronous regime (inspired by \cite{tian_asynchronous_2020}) fully decouples the communication and optimization processes.
At any given time, an agent may optimize locally with the latest information it has.
We assume that all agents will eventually be able to connect to other agents over the network, possibly after some amount of delay due to network latency or dropout.
Thus, any robot may be performing local optimization with stale data from its neighbors.
This paradigm mimics a real-world implementation with delays.

We follow \cite{tian_asynchronous_2020} and implement asynchronous \name{} by separating the communication and optimization tasks into parallel software processes.
When connected to neighboring robots, the communication process sends the overlapped portions of its latest local estimate \Vb{} to those robots (Step~\ref{alg:robo_share}) and handles incoming updates.
This process receives and caches neighbor state updates for the optimization process to access in Step~\ref{alg:robo_update}.
The optimization process meanwhile operates in a loop in which it 1) updates the local state, \Vbwplus using the cached updates from the communication process; 2) solves~\eqref{eqn:pgo_subproblem} over \Vbw{}; and 3) provides the latest $V_{\alpha}$ back to the communication process.
As in \cite{tian_asynchronous_2020}, we model the individual robots' choice of when to optimize as a Poisson process with rate $\lambda > 0$, i.e., the times $t_k \geq 0 $ between successive optimizations follow an exponential distribution, $P(t_k \leq a) = 1 - e^{-\lambda a}$ for $a \geq 0$.
All agents share the same $\lambda$, which is set according to the robots' local computation capacity and the network capacity.
Note that both processes must access the shared cache containing the latest local estimate for \Vb{} and the latest neighbor states.
Each access must be done atomically to maintain data integrity.
In practice, this can be easily implemented using software locks on the cache.

\section{Experiments}\label{sec:experiments}

We examine the performance of \name{} with different levels of overlap under the three communication paradigms described in Section~\ref{sec:methods}.
We use chordal distance as our distance metric unless explicitly noted.
For all experiments below, we use a collection of 6 simulated and 16 real-world benchmark PGO datasets\footnote{https://github.com/mit-acl/dpgo/tree/main/data}$^,$\footnote{https://github.com/MurpheyLab/DPGO/tree/main/dataset}.
This collection consists of 15 datasets with 2D measurements and 7 with 3D measurements.
For the \texttt{garage} dataset, we use noise standard deviations of 0.4 deg and 0.04 m to avoid numerical instabilities caused by the unrealistically large uncertainties ($> 40$~deg) in the original dataset.
For each benchmark, we use SE-Sync \cite{rosen_se_sync_2019} to compute a reference solution and optimal cost for the centralized PGO problem.
When using the geodesic distance metric, we find a reference solution by solving the centralized problem using Ceres \cite{Agarwal_Ceres_Solver_2022} starting from chordal initialization.
We show results only up to $\omega = 3$ as we observe diminishing returns in convergence improvement for $\omega > 3$. 

The benchmarks used are single-agent datasets, so we partition them to simulate multiple robots by segmenting the datasets sequentially into subgraphs according to the pose numbering provided within each dataset.
We simulate 5 robots for all benchmarks.
When available, we compare our method against state-of-the-art baseline methods that use the same communication paradigm and distance metric.
For the methods we compare against, we use the implementations provided by the authors, unmodified except where noted.

\subsection{Evaluation Metrics}\label{sec:metrics}

We consider two platform-agnostic metrics: RMSE absolute position error (APE) \cite{evo} and \textit{relative suboptimality}. We define relative suboptimality at iteration $k$ as
$    \Delta_{rel} = (F_k - F^*)/F^*$.
$F_k$ is the value of the cost in~\eqref{eqn:pgo} evaluated at iteration $k$, and $F^*$ is the optimal cost for the given graph. To evaluate $F_k$ in a distributed setting, we aggregate the agents' solutions using natural ownership of each pose to compute a centralized $F_k$ at each iteration; e.g., we use robot $\alpha$'s solutions for all of robot $\alpha$'s poses, even if robot $\beta$ is locally estimating part of robot $\alpha$'s trajectory.
We use the same aggregated solution to compute APE, using the optimal solution found with either SE-Sync or Ceres (depending on distance metric used) as the reference trajectory. 

\subsection{Synchronous \name}

Synchronous \name{} follows Algorithm~\ref{alg:robo} with $\mathcal{B}_{\mathrm{conn}} = \mathcal{B}$. \namew{$\omega$} denotes \name{} with a specific overlap size, e.g., \namew{2}.
For all methods, we limit each experiment to 1000 iterations. One iteration is all selected agents performing one round of communication and optimization. 

\subsubsection{Chordal Cost Function}

For the chordal distance PGO formulation, we compare \name{} with different levels of overlap against \amm{} \cite{fan_majorization_2024} and \rbcdpp{} \cite{tian_distributed_2021}, both state-of-the-art parallel, synchronous, and fully distributed approaches to PGO that also use the chordal distance objective.
We run \rbcdpp{} with greedy block selection and a fixed restart frequency of 30 iterations. We expect \amm{} to outperform \rbcdpp{} per \cite{fan_majorization_2024}, but compare against \rbcdpp{} as \name{} is also based on RBCD (without acceleration \cite{tian_distributed_2021}).

We assess the performance of these methods using three initialization scenarios: \emph{distributed chordal}, \emph{block-wise spanning tree}, and \emph{odometry}.
These schemes are presented in increasing order of difficulty in terms of the resulting optimization, and decreasing order of complexity in terms of communication required.
The selected techniques can all be performed without a central node.
We use the distributed chordal initialization approach from \cite{fan_majorization_2020} and the block-wise spanning tree method from \cite{tian_spectral_2024}. 
In the third technique, each agent initializes from its local odometry measurements only.

\begin{table}[t]
	\setlength{\tabcolsep}{1.5pt}
	\renewcommand{\arraystretch}{1.2}
	\centering
	\caption{Synchronous \name{} and baseline methods on benchmark datasets with varied initialization. Percentage of problems solved shown after 100, 500, 1000 iterations. Colors show best (green) and second-best (yellow) results.}
    \vspace*{-0.05in}
	{
    \begin{tabular}{|c||c|c|c||c|c|c||c|c|c|}
    \hline
    \multirow{3}{*}{\textbf{Method}} & \multicolumn{9}{c|}{\textbf{\% Problems Solved at Iteration ($\Delta_{rel} \leq 0.1\%$)}} \\ 
    \cline{2-10} 
     & \multicolumn{3}{c||}{Dist. Chordal} & \multicolumn{3}{c||}{Spanning Tree} & \multicolumn{3}{c|}{Odometry} \\ 
    \cline{2-10} 
    \cline{2-10}
    & 100 & 500 & 1000 & 100 & 500 & 1000 & 100 & 500 & 1000 \\
    \hline \hline

    \rbcdpp & 45.5 & 45.5 & 59.1 & 13.6 & 27.3 & 31.8 & 0.00 & 9.1 & 22.7 \\
    \hline

    \amm & 40.9 & \cellcolor{lightyellow} 81.8 & \cellcolor{lightgreen} \textbf{90.9} & \cellcolor{lightyellow} 18.2 & \cellcolor{lightyellow} 54.5 & 63.6 & 0.00 & 22.7 & 54.5 \\
    \hline

    \namew{1} & 45.5 & 72.7 & 81.8 & 9.1 & 40.9 & 45.5 & \cellcolor{lightyellow} 9.1 & 18.2 & 27.3 \\
    \hline

    \namew{2} & \cellcolor{lightyellow} 63.6 & \cellcolor{lightyellow} 81.8 & \cellcolor{lightyellow} 86.4 & 13.6 & \cellcolor{lightyellow} 54.5 & \cellcolor{lightyellow} 72.7 & \cellcolor{lightgreen} \textbf{13.6} & \cellcolor{lightyellow} 40.9 & \cellcolor{lightgreen} \textbf{63.6} \\
    \hline

    \namew{3} & \cellcolor{lightgreen} \textbf{72.7} & \cellcolor{lightgreen} \textbf{86.4} & \cellcolor{lightgreen} \textbf{90.9} & \cellcolor{lightgreen} \textbf{31.8} & \cellcolor{lightgreen} \textbf{77.3} & \cellcolor{lightgreen} \textbf{86.4} & \cellcolor{lightgreen} \textbf{13.6} & \cellcolor{lightgreen} \textbf{50.0} & \cellcolor{lightyellow} 59.1 \\
    \hline

	\end{tabular}}
	\label{tab:parallel_chordal_results}
    \vspace*{-0.1in}
\end{table}

\begin{figure}[t]
    \centering
    \includegraphics[width=1.0\linewidth]{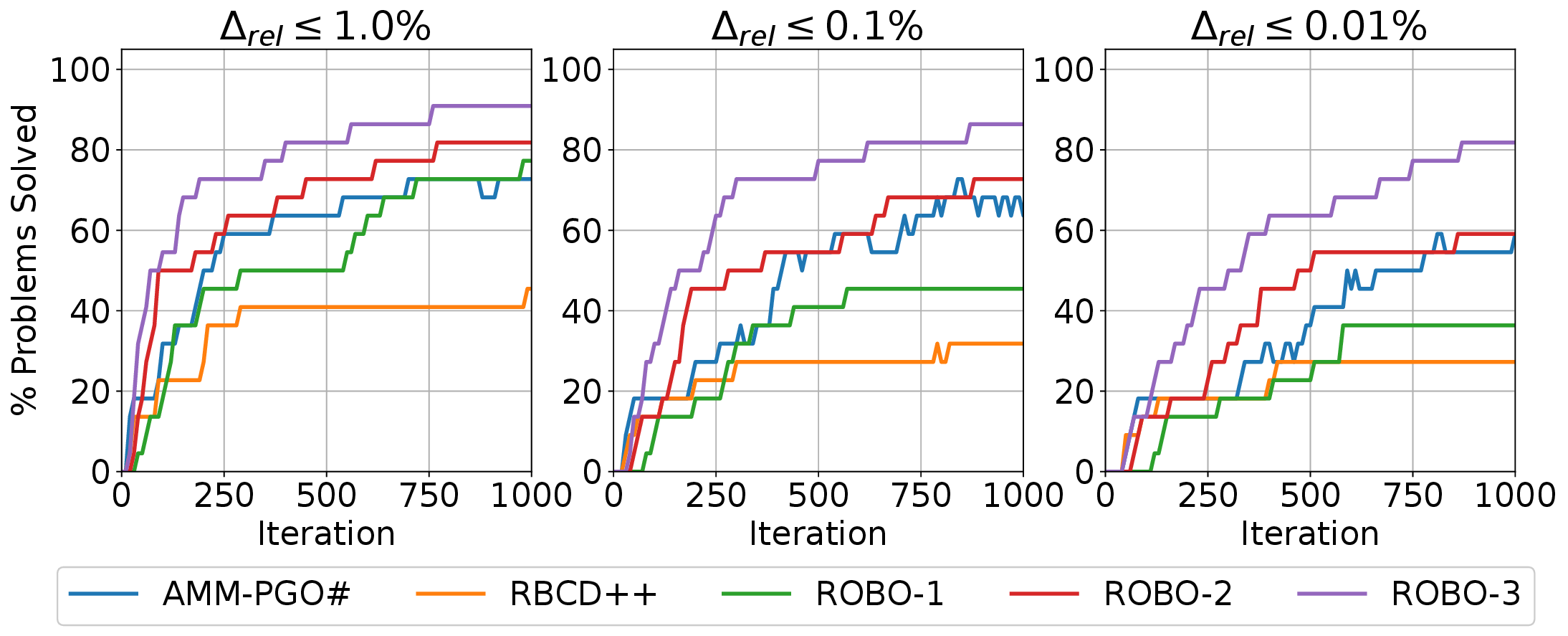}
    \caption{Performance profile comparing \amm, \rbcdpp{}, and \name{} with varying overlap levels when using block-wise spanning tree initialization. From left to right, the relative suboptimality thresholds used to determine convergence are 1\%, 0.1\%, and 0.01\%.}
    \label{fig:block_performance} \vspace*{-0.28in}
\end{figure}

The results of all three initialization scenarios for a relative suboptimality threshold \relsubopt{} of 0.1\% are summarized in Table~\ref{tab:parallel_chordal_results}.
We consider the percentage of datasets on which each method has converged after 100, 500, and 1000 iterations.
In all experiments shown, \namew{$\omega$} for any $\omega$ and \amm{} reach convergence on more of the benchmark datasets within 1000 iterations than \rbcdpp{}.
With distributed chordal initialization, 
\namew{2} and \namew{3} perform on par with \amm{}.
But for spanning tree and odometry initialization, which are easier to implement but result in a harder distributed optimization, \namew{2} and \namew{3} outperform both baselines.
We also create a performance profile as in \cite{fan_majorization_2024} showing the number of iterations required to reach a particular \relsubopt.
Figure~\ref{fig:block_performance} shows the profile corresponding to block-wise spanning tree initialization for $\Delta_{rel} = 1\%,\ 0.1\%,\ 0.01\%$, from left to right.
We note that due to the restart mechanism in both baseline methods, their corresponding performance lines can show oscillations.
Figure~\ref{fig:block_performance} further highlights that for tighter \relsubopt{} thresholds, \namew{2} and \namew{3} also outperform both baselines.
This trend is also present with odometry initialization, though we only include one profile here due to space limitations.

\begin{table}[t]
	\setlength{\tabcolsep}{1.5pt}
	\renewcommand{\arraystretch}{1.2}
	\centering
	\caption{Cost after 100 iterations of \name{} and best-performing baseline in synchronous setting.}
    \vspace*{-0.05in}
	{
    \begin{tabular}{|c||c|c||c|c|c|}
    \hline
    \multirow{2}{*}{\textbf{Dataset}} & \multirow{2}{*}{\textbf{Opt.}} & \multirow{2}{*}{\textbf{Init.}} & \multicolumn{3}{c|}{\textbf{Objective after 100 iterations}} \\ 
    \cline{4-6} 
     & & & \amm & \namew{2} & \namew{3} \\ 
    \hline \hline

    ais2klinik (2D) & 188.5 & 2.6e8 & \cellcolor{lightgreen} \textbf{488} & \cellcolor{lightyellow} 1742 & 3.4e4 \\
    \hline

    city (2D) & 638.6 & 1.1e7 & \cellcolor{lightyellow} 717.8 & \cellcolor{lightgreen} \textbf{638.6} & \cellcolor{lightgreen} \textbf{638.6} \\
    \hline

    cubicle (3D) & 717.1 & 2.5e6 & \cellcolor{lightyellow} 852.9 & \cellcolor{lightgreen} \textbf{717.1} & \cellcolor{lightgreen} \textbf{717.1} \\
    \hline

    grid (3D) & 84319.01 & 4.9e5 & \cellcolor{lightyellow} 84319.10 & \cellcolor{lightgreen} \textbf{84319.01} & \cellcolor{lightgreen} \textbf{84319.01} \\
    \hline

    rim (3D) & 5460.9 & 3.8e7 & 8702.3 & \cellcolor{lightyellow} 5461.7 & \cellcolor{lightgreen} \textbf{5460.9} \\
    \hline

    torus (3D) & 24227.05 & 6.3e4 & \cellcolor{lightyellow} 24227.37 & \cellcolor{lightgreen} \textbf{24227.05} & \cellcolor{lightgreen} \textbf{24227.05} \\
    \hline

	\end{tabular}}
	\label{tab:dataset_results}
    \vspace*{-0.24in}
\end{table}

Table~\ref{tab:dataset_results} enumerates the cost function value after 100 iterations for the three best-performing methods on individual datasets.
Due to space constraints, we include only datasets with at least 5000 poses.
Optimal costs are calculated using SE-Sync \cite{rosen_se_sync_2019}, and initial costs come from block-wise spanning tree initialization.
We note that with this initialization, \name{} falls into local minima on \texttt{ais2klinik}, whereas our method converges on the same dataset with distributed chordal initialization.
On all other large datasets, \namew{2} and \namew{3} perform better than the baselines.

\begin{figure}[t]
    \centering
    \subfloat[t][]{\includegraphics[width=0.5\linewidth]{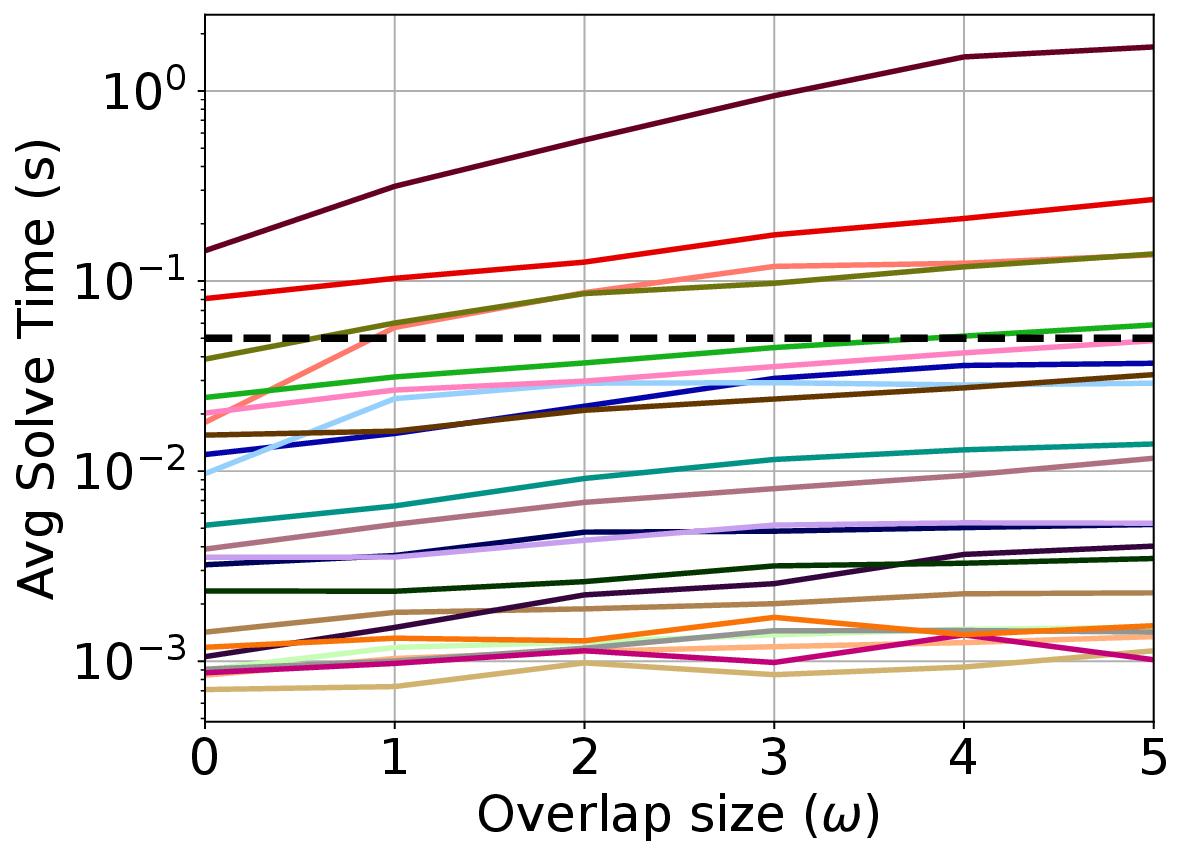}\label{fig:computation}}
    \subfloat[t][]{%
        \adjustbox{width=0.44\linewidth,valign=B,raise=1.0\baselineskip}{%
        \renewcommand{\arraystretch}{1.5}%
        \footnotesize
        \begin{tabular}{|c||c|c|}
            \hline
            \textbf{Dataset} & \textbf{Poses} & \textbf{Edges}  \\
            \hline
            grid (3D) & 8000 & 22236 \\
            \hline
            rim (3D) & 10195 & 29743 \\
            \hline
            cubicle (3D) & 5750 & 16727 \\
            \hline
            city (2D) & 10000 & 20687 \\
            \hline
            torus (3D) & 5000 & 9048 \\
            \hline
        \end{tabular}}
        \label{tab:computation}}
    \caption{(a) Local solve time as $\omega$ increases. Each curve shown is one dataset. For all but 4 of 22 datasets, local solve times are $\geq 20$~Hz (dashed line) for $\omega \leq 3$. (b) Datasets with longest solve times (decreasing order).}
    \vspace*{-0.28in}
\end{figure}

\begin{table}[t]
	\setlength{\tabcolsep}{1.5pt}
	\renewcommand{\arraystretch}{1.0}
	\centering
	\caption{ Speedup and robot-to-robot communication load of \namew{$\omega$} vs. \amm across all benchmarks.}
	{
    \begin{tabular}{|l||c|c|c|}
    \hline
     & \textbf{\amm} & \textbf{\namew{2}} & \textbf{\namew{3}} \\
    \hline \hline

    Avg. poses per iteration & 50 & 130 & 159 \\
    \hline
    Max. poses per iteration & 481 & 1010 & 1153 \\
    \hline
    Avg. data per iteration (Kb) & 11 & 29 & 36 \\
    \hline
    Max. data per iteration (Kb) & 108 & 226 & 258 \\
    \hline
    Avg. speedup over \amm{} & - & \textbf{2.6} & \textbf{3.1} \\
    \hline

	\end{tabular}}
	\label{tab:parallel_comms}
    \vspace*{-0.07in}
\end{table}

\subsubsection{Computation and Communication} 
The above results show that, in order to achieve faster convergence than state-of-the-art methods on benchmark datasets, \name{} requires $\omega \geq 2$.
To assess the feasibility of using overlapping blocks of this size, we first consider the computation cost of locally solving larger subproblems.
Figure~\ref{fig:computation} shows the per-robot, single iteration computation time increase on all datasets with respect to $\omega$ as run on an Intel Xeon E-2276M 2.8 GHz processor with 32 GB of RAM.
We observe the relative solve times between datasets, and the relative change in computation as $\omega$ increases, as these are primarily a function of graph structure and will be platform- and implementation-independent.
The datasets with the five largest solve times are listed in Figure~\ref{tab:computation}.
These datasets have larger graphs ($\geq 5000$ poses) and are highly connected (i.e., many more edges than poses), which leads to greater relative increase in the domain size with larger $\omega$ than on a sparser graph.
For the majority of the benchmarks, we observe mild or moderate increase in local solve time, demonstrating the practicality of \name{} when local computation is not extremely limited.

We next examine the robot-to-robot communication per iteration of \name{} relative to \amm.
We calculate the data cost per iteration by counting the number of poses shared between each pair of robots, including prior poses.
We consider both the average and maximum number of poses shared between all pairs of neighboring robots at each iteration.
Table~\ref{tab:parallel_comms} shows this average and maximum over all benchmark datasets, as well as the amount of data shared if each pose is represented by seven 32-bit floating point values.
We also compare the number of iterations required to converge to $\Delta_{rel}= 0.1\%$ 
from distributed chordal initialization, noting ``speedup'' as a multiplier compared to the baseline method \amm{}.
We see that \namew{2} and \namew{3} offer an average speedup of $\mathbf{2.6\times}$ and $\mathbf{3.1\times}$ over \amm{} with an average of \textbf{only 25 Kb} additional per-iteration communication cost between robot pairs.
In the worst case (corresponding to the \texttt{city} dataset), \name{} doubles the total network communication cost to roughly 250 Kb. 
Even with 100 iterations per second, this case would require 25 Mbps of bandwidth (versus the baseline of about 11 Mbps for the same iteration rate)---well within the limits of a 100 Mbps radio network like the Silvus SL5200 \cite{silvus}.
Thus, considering both computation and communication, we conclude that, for high-bandwidth wireless networks often used in ground and air robotics applications, the additional data cost of \name{} for $\omega \leq 3$ is minor, even for large pose graphs. 
In practice, $\omega$ can be set to maximize available computation and communication. 
However, as previously noted, $\omega > 3$ leads to diminishing returns, which should be considered when selecting $\omega$ for a particular system.

\subsubsection{Geodesic Cost Function}

We compare \name{} with a geodesic distance cost against MESA \cite{mcgann_asynchronous_2024}, a state-of-the-art C-ADMM approach to DPGO that also uses the geodesic distance in its cost formulation \cite{gtsam}.
For MESA, we use the hyperparameters the authors provide for benchmark dataset experiments.
To match our parallel communication paradigm, we modify the original implementation of MESA\footnote{https://github.com/rpl-cmu/mesa} to allow all robots to optimize at every iteration in a synchronized fashion, which we denote as MESA-parallel.
We use distributed chordal initialization in these comparisons.
As tuning of the hyperparameters for MESA may affect the rate of decrease in the optimality gap, we also assess the RMSE APE.
Once the RMSE APE is $\leq 0.1\ \mathrm{m}$, we consider that dataset solved.
Table~\ref{tab:all_geodesic_results} (top) shows these summarized results for MESA-parallel and \namew{$\omega$} with a geodesic distance cost function after 100, 500, and 1000 iterations.
While \name{} with geodesic cost converges more slowly than with chordal cost, \name{} with any overlap converges more quickly than MESA-parallel in terms of both metrics.

\begin{table}[t]
	\setlength{\tabcolsep}{1.5pt}
	\renewcommand{\arraystretch}{1.2}
	\centering
	\caption{\name{} (geodesic cost) vs MESA-parallel and MESA}
    \vspace*{-0.05in}
	{
    \begin{tabular}{c|c||c|c|c||c|c|c|}
    \cline{2-8}
    & \multirow{3}{*}{\textbf{Method}} & \multicolumn{6}{c|}{\textbf{\% Problems Solved at Iteration}} \\ 
    \cline{3-8} 
     & & \multicolumn{3}{c||}{\relsubopt{} $\leq 0.1\%$} & \multicolumn{3}{c|}{RMSE APE $\leq 0.1$} \\ 
    \cline{3-8} 
    \cline{3-8}
    & & 100 & 500 & 1000 & 100 & 500 & 1000 \\
    \cline{2-8}
    \hline

    \multirow{4}{*}{\rotatebox{90}{\textbf{Sync.}}}
    & MESA-parallel & 13.6 & 27.3 & 27.3 & 31.8 & 31.8 & 31.8 \\
    \hhline{~|-------|}

    & \namew{1} & 45.5 & 59.1 & 72.7 & 45.5 & 54.5 & \cellcolor{lightyellow} 59.1 \\
    \hhline{~|-------|}

    & \namew{2} & \cellcolor{lightyellow} 50.0 & \cellcolor{lightyellow} 68.2 & \cellcolor{lightyellow} 77.3 & \cellcolor{lightyellow} 54.5 & \cellcolor{lightyellow} 63.6 & \cellcolor{lightgreen} \textbf{72.7} \\
    \hhline{~|-------|}

    & \namew{3} & \cellcolor{lightgreen} \textbf{59.1} & \cellcolor{lightgreen} \textbf{72.7} & \cellcolor{lightgreen} \textbf{81.8} & \cellcolor{lightgreen} \textbf{59.1} & \cellcolor{lightgreen} \textbf{68.2} & \cellcolor{lightgreen} \textbf{72.7} \\
    \hline

    \hline
    \multirow{4}{*}{\rotatebox{90}{\textbf{Edgewise}}}
    & MESA & 0.0 & 18.2 & 31.8 & 27.3 & 31.8 & 31.8 \\
    \hhline{~|-------|}

    & \namew{1} & \cellcolor{lightyellow} 31.8 & 40.9 & 45.5 & 40.9 & 45.5 & 45.5 \\
    \hhline{~|-------|}

    & \namew{2} & \cellcolor{lightyellow} 31.8 & \cellcolor{lightyellow} 45.5 & \cellcolor{lightyellow} 54.5 & \cellcolor{lightyellow} 45.5 & \cellcolor{lightyellow} 54.5 & \cellcolor{lightyellow} 54.5 \\
    \hhline{~|-------|}

    & \namew{3} & \cellcolor{lightgreen} \textbf{40.9} & \cellcolor{lightgreen} \textbf{59.1} & \cellcolor{lightgreen} \textbf{63.6} & \cellcolor{lightgreen} \textbf{50.0} & \cellcolor{lightgreen} \textbf{59.1} & \cellcolor{lightgreen} \textbf{59.1} \\
    \hline

	\end{tabular}}
	\label{tab:all_geodesic_results} \vspace*{-0.25in}
\end{table}

\subsection{Edgewise \name}

The edgewise-only paradigm restricts communication at each iteration to only one pair of connected robots (see Section~\ref{sec:methods}).
We again use the geodesic distance cost function and distributed chordal initialization.
We compare against the original implementation of  MESA \cite{mcgann_asynchronous_2024} with the same hyperparameters as in the previous experiments.
The experimental results summarized in 
Table~\ref{tab:all_geodesic_results} (bottom) show that 
both in terms of relative suboptimality and RMSE APE, \name{} converges on more benchmarks than MESA within 1000 iterations. 
We see that as in the synchronized case, in this communicated-limited setting, sharing moderately more information between agents may significantly improve convergence rates.

\subsection{Asynchronous \name}
\begin{table}[t]
	\setlength{\tabcolsep}{1.5pt}
	\renewcommand{\arraystretch}{1.2}
	\centering
	\caption{Asynchronous \name{} cost (1s delay, $\lambda = 10$Hz)}
    \vspace*{-0.05in}
	{
    \begin{tabular}{|c||c|c||c|c|c|c|c|}
    \hline
    \multirow{2}{*}{\textbf{Dataset}} & \multirow{2}{*}{\textbf{Opt.}} & \multirow{2}{*}{\textbf{Init.}} & \multicolumn{5}{c|}{\textbf{\namew{$\omega$} objective after 30~s}} \\ 
    \cline{4-8} 
     & & & $\omega=1$ & 2 & 3 & 4 & 5 \\ 
    \hline \hline

    ais2klinik (2D) & 188.5 & 322 & 196.9 & 195.5 & 194.3 & \cellcolor{lightyellow} 193.8 & \cellcolor{lightgreen} \textbf{192.6} \\
    \hline

    city (2D) & 638.6 & 711 & 650.3 & 642.6 & 640.1 & \cellcolor{lightyellow} 639.3 & \cellcolor{lightgreen} \textbf{638.8} \\
    \hline

    cubicle (3D) & 717.1 & 835 & 720.2 & 719.0 & \cellcolor{lightyellow} 718.0 & 749.5 & \cellcolor{lightgreen} \textbf{717.2} \\
    \hline

    grid (3D) & 84319 & 87265 & \cellcolor{lightgreen} \textbf{84319} & 84334 & \cellcolor{lightyellow} 84324 & \cellcolor{lightyellow} 84324 & 84436 \\
    \hline

    rim (3D) & 5460.9 & 8084 & 5890.2 & 5535.2 & 5469.3 & \cellcolor{lightyellow} 5467.1 & \cellcolor{lightgreen} \textbf{5464.6} \\
    \hline

    torus (3D) & 24227 & 24668 & \cellcolor{lightyellow} 24228 & 24230 & \cellcolor{lightgreen} \textbf{24227} & \cellcolor{lightgreen} \textbf{24227} & \cellcolor{lightgreen} \textbf{24227} \\
    \hline

	\end{tabular}}
	\label{tab:async_results}\vspace*{-0.2in}
\end{table}

\begin{figure}[t]
    \centering
    \subfloat[]{\includegraphics[width=0.49\linewidth]{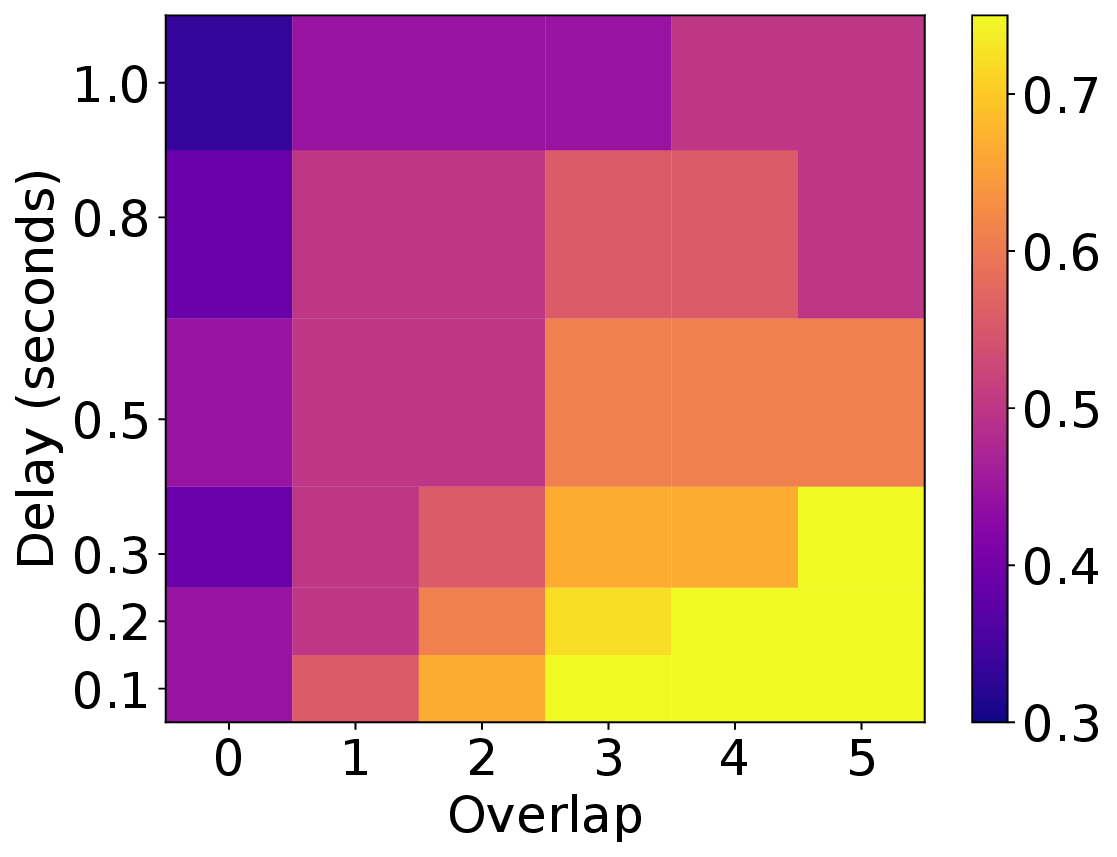}\label{fig:async_shortdelays}}
    \subfloat[]{\includegraphics[width=0.49\linewidth]{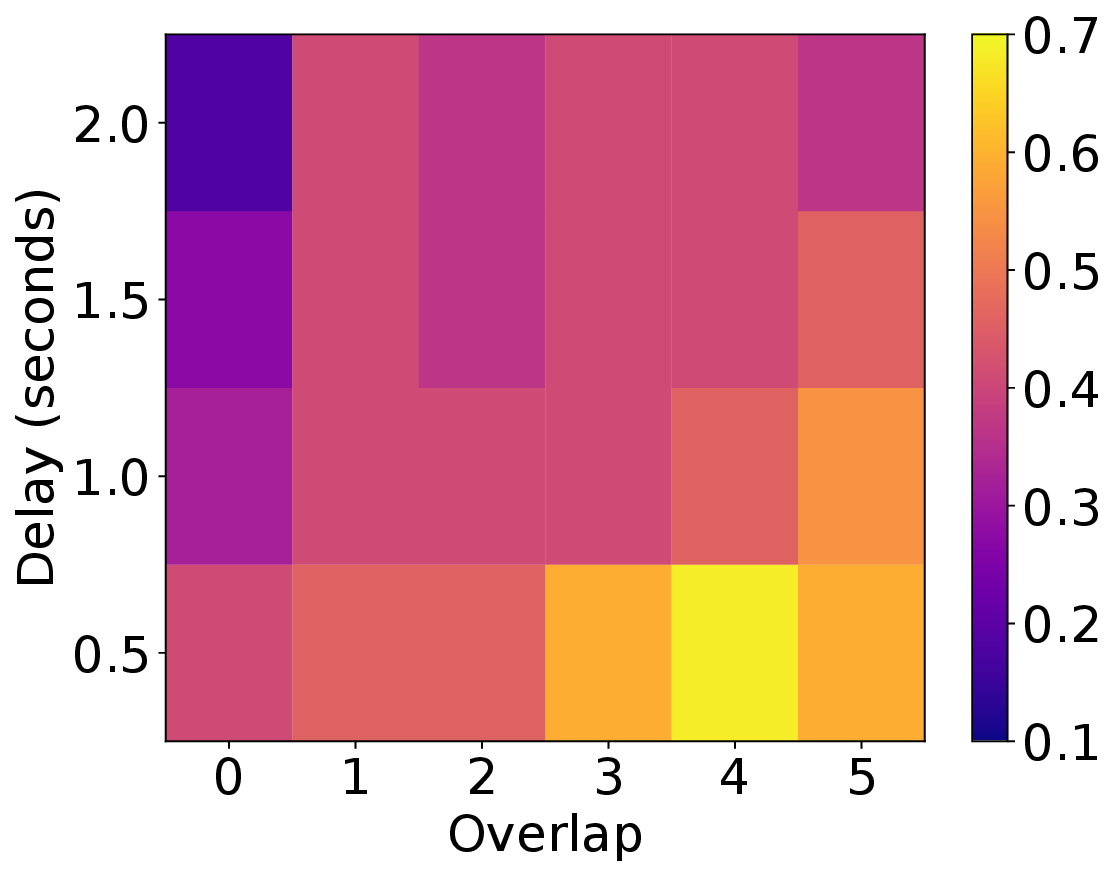}\label{fig:async_longdelays}}
    \caption{Fraction of problems solved with \relsubopt $\leq 0.1 \%$ after 30 s with increasing delay and overlap ($\lambda = 10$ Hz). (a) Shorter delays, excluding 4 datasets with longest local solve times. (b) Longer delays, all datasets.}\label{fig:async_heatmaps}
    \vspace*{-0.25in}
\end{figure}

Finally, we consider the asynchronous setting. 
In our experiments, we assess the sensitivity of \name{} to increasing delay $d$ and overlap size $\omega$.
Each agent runs as its own process on a CPU.
We use distributed chordal initialization before running all agents asynchronously for 30 seconds. 
We choose $\lambda$ such that we can isolate the impact of the delays $d$ on convergence.
The solve times in Figure~\ref{fig:computation} show that for all but 4 datasets, $\lambda=10$~Hz ensures that the expected value of the time between local optimizations is much larger than the local solve time.
This means that for these 18 datasets, we can expect local optimization to run at approximately the rate $\lambda$. 
On this subset, we can test the impacts of shorter delays ($d=0.1\mathrm{-}1$~s) without local computation time adding to the delay.
Figure~\ref{fig:async_shortdelays} summarizes the fraction of these 18 benchmarks solved after 30~s for $\Delta_{rel} \leq 0.1\%$.
Broadly, these results show that longer delays slow down convergence, and larger $\omega$ speeds up convergence for the same delay. 

In Figure~\ref{fig:async_longdelays}, we repeat our experiments with all 22 datasets and $d \geq 0.5$~s.
We choose larger jumps between delays to separate the effects of longer computation time on the largest datasets from increasing simulated delay.
As in the synchronous setting, both heatmaps in Figure~\ref{fig:async_heatmaps} show diminishing returns with increasing $\omega$, with longer delays requiring greater increase in $\omega$ to achieve improved convergence rates. 
Additionally, for delays longer than 1~s, the convergence of \name{} is more sensitive to increasing overlap. 
Table~\ref{tab:async_results} lists the objective value for datasets with $\geq 5000$ poses after 30~s of asynchronous optimization with a communication delay of 1~s. 
We note that \texttt{grid}, which has the longest local solve times for all $\omega$, experiences greater delay than the simulated 1~s as $\omega$ increases, leading also to higher costs.
For the remaining datasets, Table~\ref{tab:async_results} highlights both the improvement from larger overlap, and the greater sensitivity of \name{} convergence when both delay and $\omega$ are large. 
These results show that \name{} with $\omega \leq 3$ is robust to moderate communication delays, and that overlap can be used to improve convergence even in asynchronous settings.

\section{Conclusion}\label{sec:conclusion}
This work presented \name{}, a novel, fully parallel distributed pose graph optimization method that utilizes available communication bandwidth for significantly improved convergence rates.
We demonstrated the applicability of our method with varying levels of overlap in different initialization scenarios; with different rotation distance metrics in the objective; and in both synchronous and asynchronous settings on a wide range of benchmark datasets.
We showed our method can achieve 3.1$\times$ faster convergence over the state-of-the-art while requiring an average of only 36~Kb in per-iteration communication.
Our future work will investigate a theoretical basis for both the convergence and limitations of our approach, explore conditions under which our method might have guaranteed convergence properties, and work toward a real-world asynchronous implementation.
We will also consider the application of overlapping domain decomposition in other DPGO approaches, such as C-ADMM.

\section*{Acknowledgment}
The authors thank Daniel McGann for valuable discussions about MESA in the development of this work.

\bibliographystyle{IEEEtran}
\bibliography{IEEEabrv,dpgo_library,additional_refs}

\begin{thebibliography}{10}
\providecommand{\url}[1]{#1}
\csname url@rmstyle\endcsname
\providecommand{\newblock}{\relax}
\providecommand{\bibinfo}[2]{#2}
\providecommand\BIBentrySTDinterwordspacing{\spaceskip=0pt\relax}
\providecommand\BIBentryALTinterwordstretchfactor{4}
\providecommand\BIBentryALTinterwordspacing{\spaceskip=\fontdimen2\font plus
\BIBentryALTinterwordstretchfactor\fontdimen3\font minus
  \fontdimen4\font\relax}
\providecommand\BIBforeignlanguage[2]{{%
\expandafter\ifx\csname l@#1\endcsname\relax
\typeout{** WARNING: IEEEtran.bst: No hyphenation pattern has been}%
\typeout{** loaded for the language `#1'. Using the pattern for}%
\typeout{** the default language instead.}%
\else
\language=\csname l@#1\endcsname
\fi
#2}}
\renewcommand\BIBentryALTinterwordstretchfactor{4}

\bibitem{ebadi_present_2024}
K.~Ebadi, L.~Bernreiter, \emph{et~al.}, ``Present and {Future} of {SLAM} in
  {Extreme} {Environments}: {The} {DARPA} {SubT} {Challenge},'' \emph{IEEE
  Trans. on Robotics}, vol.~40, pp. 936--959, 2024.

\bibitem{chang_kimera-multi_2021}
Y.~Chang, Y.~Tian, J.~P. How, and L.~Carlone, ``Kimera-{Multi}: a {System} for
  {Distributed} {Multi}-{Robot} {Metric}-{Semantic} {Simultaneous}
  {Localization} and {Mapping},'' in \emph{2021 {IEEE} {Int.} {Conf.} on
  {Robotics} and {Automation}}, 2021, pp. 11\,210--11\,218.

\bibitem{lajoie_door-slam_2020}
P.-Y. Lajoie, B.~Ramtoula, \emph{et~al.}, ``{DOOR}-{SLAM}: {Distributed},
  {Online}, and {Outlier} {Resilient} {SLAM} for {Robotic} {Teams},''
  \emph{IEEE Robotics and Automation Letters}, vol. 5(2), pp. 1656--1663, Apr.
  2020.

\bibitem{chang_lamp_2022}
Y.~Chang, K.~Ebadi, \emph{et~al.}, ``{LAMP} 2.0: {A} {Robust} {Multi}-{Robot}
  {SLAM} {System} for {Operation} in {Challenging} {Large}-{Scale}
  {Underground} {Environments},'' \emph{IEEE Robotics and Automation Letters},
  vol. 7(4), pp. 9175--9182, Oct. 2022.

\bibitem{schmuck_covins_2021}
P.~Schmuck, T.~Ziegler, \emph{et~al.}, ``{COVINS}: {Visual}-{Inertial} {SLAM}
  for {Centralized} {Collaboration},'' in \emph{2021 {IEEE} {Int.} {Symposium}
  on {Mixed} and {Augmented} {Reality} {Adjunct}}, Oct. 2021, pp. 171--176.

\bibitem{li_corb-slam_2018}
F.~Li, S.~Yang, X.~Yi, and X.~Yang, ``\BIBforeignlanguage{en}{{CORB}-{SLAM}:
  {A} {Collaborative} {Visual} {SLAM} {System} for {Multiple} {Robots}},'' in
  \emph{\BIBforeignlanguage{en}{Collaborative {Computing}: {Networking},
  {Applications} and {Worksharing}}}.\hskip 1em plus 0.5em minus 0.4em\relax
  Springer International Publishing, 2018, pp. 480--490.

\bibitem{gtsam}
\BIBentryALTinterwordspacing
F.~Dellaert and {GTSAM Contributors}, ``borglab/gtsam,'' May 2022. [Online].
  Available: \url{https://github.com/borglab/gtsam}
\BIBentrySTDinterwordspacing

\bibitem{Agarwal_Ceres_Solver_2022}
\BIBentryALTinterwordspacing
S.~Agarwal, K.~Mierle, and {The Ceres Solver Team}, ``{Ceres Solver},'' Oct.
  2023. [Online]. Available: \url{https://github.com/ceres-solver/ceres-solver}
\BIBentrySTDinterwordspacing

\bibitem{choudhary_distributed_2017}
S.~Choudhary, L.~Carlone, \emph{et~al.}, ``\BIBforeignlanguage{en}{Distributed
  mapping with privacy and communication constraints: {Lightweight} algorithms
  and object-based models},'' \emph{\BIBforeignlanguage{en}{The Int. Journal of
  Robotics Research}}, vol. 36(12), pp. 1286--1311, Oct. 2017.

\bibitem{murai_robot_2024}
R.~Murai, J.~Ortiz, \emph{et~al.}, ``A {Robot} {Web} for {Distributed}
  {Many}-{Device} {Localization},'' \emph{IEEE Trans. on Robotics}, vol.~40,
  pp. 121--138, 2024.

\bibitem{tian_asynchronous_2020}
Y.~Tian, A.~Koppel, A.~S. Bedi, and J.~P. How, ``Asynchronous and {Parallel}
  {Distributed} {Pose} {Graph} {Optimization},'' \emph{IEEE Robotics and
  Automation Letters}, vol. 5(4), pp. 5819--5826, Oct. 2020.

\bibitem{tian_distributed_2021}
Y.~Tian, K.~Khosoussi, D.~M. Rosen, and J.~P. How, ``Distributed {Certifiably}
  {Correct} {Pose}-{Graph} {Optimization},'' \emph{IEEE Trans. on Robotics},
  vol. 37(6), pp. 2137--2156, Dec. 2021.

\bibitem{fan_majorization_2024}
T.~Fan and T.~D. Murphey, ``Majorization {Minimization} {Methods} for
  {Distributed} {Pose} {Graph} {Optimization},'' \emph{IEEE Trans.\ on
  Robotics}, vol.~40, pp. 22--42, 2024.

\bibitem{mcgann_asynchronous_2024}
D.~McGann, K.~Lassak, and M.~Kaess, ``\BIBforeignlanguage{en}{Asynchronous
  {Distributed} {Smoothing} and {Mapping} via {On}-{Manifold} {Consensus}
  {ADMM}},'' in \emph{\BIBforeignlanguage{en}{IEEE Int. Conf. on Robotics and
  Automation}}, 2024, pp. 4577--4583.

\bibitem{xbee}
\BIBentryALTinterwordspacing
\emph{{Digi} {XBee} 3 {Zigbee} 3.0 {Datasheet}}, Digi, 2023. [Online].
  Available:
  \url{https://www.digi.com/resources/library/data-sheets/ds_xbee-3-zigbee-3}
\BIBentrySTDinterwordspacing

\bibitem{silvus}
\BIBentryALTinterwordspacing
\emph{{StreamCaster} {LITE} 5200 {Data} {Sheet}}, Silvus Technologies, 2024.
  [Online]. Available:
  \url{https://silvustechnologies.com/wp-content/uploads/2024/10/StreamCaster-Lite-5200-SL5200-OEM-Module-Datasheet.pdf}
\BIBentrySTDinterwordspacing

\bibitem{saad_iterative_2007}
Y.~Saad, \emph{\BIBforeignlanguage{Eng}{Iterative {Methods} for {Sparse}
  {Linear} {Systems}}}, 2nd~ed.\hskip 1em plus 0.5em minus 0.4em\relax Society
  for Industrial and Applied Mathematics, 2007.

\bibitem{shin_decentralized_2020}
S.~Shin, V.~Zavala, and M.~Anitescu, ``Decentralized {Schemes} with {Overlap}
  for {Solving} {Graph}-{Structured} {Optimization} {Problems},'' \emph{IEEE
  Trans.\ on Control of Network Systems}, vol. 7(3), pp. 1225--1236, Sept.
  2020.

\bibitem{shin_overlapping_2020}
S.~Shin, M.~Anitescu, and V.~Zavala, ``Overlapping {Schwarz} {Decomposition}
  for {Constrained} {Quadratic} {Programs},'' in \emph{2020 59th {IEEE} {Conf.}
  on {Decision} and {Control}}, 2020, pp. 3004--3009.

\bibitem{shin_exponential_2021}
------, ``Exponential {Decay} of {Sensitivity} in {Graph}-{Structured}
  {Nonlinear} {Programs},'' Dec. 2021, arXiv:2101.03067.

\bibitem{andersson_c-sam_2008}
L.~A.~A. Andersson and J.~Nygards, ``\BIBforeignlanguage{en}{C-{SAM}:
  {Multi}-{Robot} {SLAM} using square root information smoothing},'' in
  \emph{\BIBforeignlanguage{en}{2008 {IEEE} {Int.} {Conf.} on {Robotics} and
  {Automation}}}, 2008, pp. 2798--2805.

\bibitem{bailey_decentralised_2011}
T.~Bailey, M.~Bryson, \emph{et~al.}, ``Decentralised cooperative localisation
  for heterogeneous teams of mobile robots,'' in \emph{2011 {IEEE} {Int.}
  {Conf.} on {Robotics} and {Automation}}, 2011, pp. 2859--2865.

\bibitem{zhang_mr-isam2_2021}
Y.~Zhang, M.~Hsiao, \emph{et~al.}, ``{MR}-{iSAM2}: {Incremental} {Smoothing}
  and {Mapping} with {Multi}-{Root} {Bayes} {Tree} for {Multi}-{Robot}
  {SLAM},'' in \emph{2021 {IEEE}/{RSJ} {Int.} {Conf.} on {Intelligent} {Robots}
  and {Systems}}, 2021, pp. 8671--8678.

\bibitem{liu_slideslam_2024}
X.~Liu, J.~Lei, \emph{et~al.}, ``{SlideSLAM}: {Sparse}, {Lightweight},
  {Decentralized} {Metric}-{Semantic} {SLAM} for {Multi}-{Robot}
  {Navigation},'' July 2024, arXiv:2406.17249.

\bibitem{lajoie_swarm-slam_2024}
P.-Y. Lajoie and G.~Beltrame, ``Swarm-{SLAM} : {Sparse} {Decentralized}
  {Collaborative} {Simultaneous} {Localization} and {Mapping} {Framework} for
  {Multi}-{Robot} {Systems},'' \emph{IEEE Robotics and Automation Letters},
  vol. 9(1), pp. 475--482, Jan. 2024.

\bibitem{cunningham_ddf-sam_2010}
A.~Cunningham, M.~Paluri, and F.~Dellaert, ``{DDF}-{SAM}: {Fully} distributed
  {SLAM} using {Constrained} {Factor} {Graphs},'' in \emph{2010 {IEEE}/{RSJ}
  {Int.} {Conf.} on {Intelligent} {Robots} and {Systems}}, 2010, pp.
  3025--3030.

\bibitem{cunningham_ddf-sam_2013}
A.~Cunningham, V.~Indelman, and F.~Dellaert, ``{DDF}-{SAM} 2.0: {Consistent}
  distributed smoothing and mapping,'' in \emph{2013 {IEEE} {Int.} {Conf.} on
  {Robotics} and {Automation}}, 2013, pp. 5220--5227.

\bibitem{halsted_survey_2021}
T.~Halsted, O.~Shorinwa, J.~Yu, and M.~Schwager, ``\BIBforeignlanguage{en}{A
  {Survey} of {Distributed} {Optimization} {Methods} for {Multi}-{Robot}
  {Systems}},'' Mar. 2021, arXiv:2103.12840.

\bibitem{fan_majorization_2020}
T.~Fan and T.~Murphey, ``\BIBforeignlanguage{en}{Majorization {Minimization}
  {Methods} for {Distributed} {Pose} {Graph} {Optimization} with {Convergence}
  {Guarantees}},'' in \emph{\BIBforeignlanguage{en}{{IEEE} {Int.} {Conf.} on
  {Int.} {Robots} and {Systems}}}, 2020, pp. 5058--5065.

\bibitem{li_distributed_2024}
C.~Li, G.~Guo, P.~Yi, and Y.~Hong, ``Distributed {Pose}-{Graph} {Optimization}
  {With} {Multi}-{Level} {Partitioning} for {Multi}-{Robot} {SLAM},''
  \emph{IEEE Robotics and Automation Letters}, vol. 9(6), pp. 4926--4933, June
  2024.

\bibitem{hamam_streaming_2022}
T.~H. Hamam and J.~Romberg, ``Streaming {Solutions} for {Time}-{Varying}
  {Optimization} {Problems},'' \emph{IEEE Trans. on Signal Processing},
  vol.~70, pp. 3582--3597, 2022.

\bibitem{na_convergence_2022}
S.~Na, S.~Shin, M.~Anitescu, and V.~Zavala, ``On the {Convergence} of
  {Overlapping} {Schwarz} {Decomposition} for {Nonlinear} {Optimal}
  {Control},'' \emph{IEEE Trans. on Auto.~Control}, vol. 67(11), pp.
  5996--6011, Mar. 2022.

\bibitem{rosen_se_sync_2019}
D.~M. Rosen, L.~Carlone, A.~S. Bandeira, and J.~J. Leonard, ``{SE-Sync}: A
  certifiably correct algorithm for synchronization over the special
  {Euclidean} group,'' \emph{The Int. Journal of Robotics Research}, vol.
  38(2-3), pp. 95--125, 2019.

\bibitem{hartley_rotation_2013}
R.~Hartley, J.~Trumpf, Y.~Dai, and H.~Li, ``\BIBforeignlanguage{en}{Rotation
  {Averaging}},'' \emph{\BIBforeignlanguage{en}{Int. Journal of Computer
  Vision}}, vol. 103, no.~3, pp. 267--305, July 2013.

\bibitem{evo}
M.~Grupp, ``evo: Python package for the evaluation of odometry and {SLAM}.''
  \url{https://github.com/MichaelGrupp/evo}, 2017.

\bibitem{tian_spectral_2024}
Y.~Tian and J.~P. How, ``Spectral {Sparsification} for
  {Communication}-{Efficient} {Collaborative} {Rotation} and {Translation}
  {Estimation},'' \emph{IEEE Trans. on Robotics}, vol.~40, pp. 257--276, 2024.

\end{thebibliography}

\end{document}